\documentclass[sigconf]{acmart}
\AtBeginDocument{%
  \providecommand\BibTeX{{%
    \normalfont B\kern-0.5em{\scshape i\kern-0.25em b}\kern-0.8em\TeX}}}

\setcopyright{acmcopyright}
\copyrightyear{2021}
\acmYear{2021}
\acmDOI{10.1145/1122445.1122456}

\acmConference[AISEC '21]{AISEC '21: ACM WORKSHOP ON ARTIFICIAL INTELLIGENCE AND SECURITY}{November 15--19, 2021}{Seoul, South Korea}
\acmBooktitle{AISEC '21: ACM WORKSHOP ON ARTIFICIAL INTELLIGENCE AND SECURITY,
  November 15--19, 2021, Seoul, South Korea}
\acmPrice{15.00}
\acmISBN{978-1-4503-XXXX-X/18/06}

\begin{document}

\title{Who's Afraid of Thomas Bayes?}

\author{Erick Galinkin}
\email{erick\_galinkin@rapid7.com}
\orcid{0000-0003-1268-9258}
\affiliation{%
  \institution{Drexel University}
  \streetaddress{3141 Chestnut St}
  \city{Philadelphia}
  \state{Pennsylvania}
  \country{USA}
  \postcode{19104}
}
\affiliation{%
  \institution{Rapid7}
  \streetaddress{120 Causeway St}
  \city{Boston}
  \state{Massachusetts}
  \country{USA}
  \postcode{02114}
}

\renewcommand{\shortauthors}{Galinkin}

\begin{abstract}
In many cases, neural networks perform well on test data, but tend to overestimate their confidence on out-of-distribution data.
This has led to adoption of Bayesian neural networks, which better capture uncertainty and therefore more accurately reflect the model's confidence.
For machine learning security researchers, this raises the natural question of how making a model Bayesian affects the security of the model.
In this work, we explore the interplay between Bayesianism and two measures of security: model privacy and adversarial robustness.
We demonstrate that Bayesian neural networks are more vulnerable to membership inference attacks in general, but are at least as robust as their non-Bayesian counterparts to adversarial examples.
\end{abstract}



\maketitle

\section{Introduction}
Deep learning has revolutionized machine learning over the last decade, providing tremendous advancements across disciplines and exploding the public's interest in artificial intelligence.
One of the safety challenges faced by deep neural networks~\cite{amodei2016concrete} has been the inability to represent uncertainty, especially on data which lies outside of the model's training distribution.
Attempts to rectify this challenge have fueled new research in the field of Bayesian neural networks, since Bayesianism reduces overconfidence~\cite{kristiadi2020being} in these models.
The research of Yarin Gal~\cite{gal2016dropout}; Wilson and Ismailov~\cite{wilson2020bayesian}; and Kristiadi \textit{et al.} has provided tools to make familiar model architectures more Bayesian, which enhances the trustworthiness and safety of neural networks.
The ease with which familiar architectures can now incorporate a Bayesian component lowers the barriers to development and deployment of Bayesian models, suggesting that we will see growing deployment of these models.

Safety in machine learning is, however, much more than simply avoiding overconfident predictions.
In order for us to ensure that machine learning systems are safe, we must also work to keep them secure.
Security-conscious AI researchers may wonder if there is some side effect to making models more Bayesian on the security and privacy of our models, a question that this work seeks to address.
Although there are other possible ways to quantify uncertainty, Bayesian neural networks have gained traction in industry as a way to alleviate this overconfidence, and have a reputation for being theoretically sound.

In this work, we wish to examine both how Bayesianism impacts the confidentiality of the model's training data and the integrity of the model's predictions.
We accomplish this by first exploring how additional information conferred by making the model more Bayesian can enable privacy-damaging attacks like membership inference~\cite{shokri2017membership} which seek to determine if a sample was part of the target model's training data.
Then, we examine how specially crafted samples that aim to cause misclassification~\cite{szegedy2013intriguing} -- so-called adversarial examples -- are affected by Bayesianism in neural networks.

\section{Background} \label{sec:background}
This work builds on the body of literature surrounding Bayesian neural networks, membership inference attacks, and adversarial machine learning.
Neural networks are machine learning models which couple an affine transformation with a nonlinear activation function and have been widely used in the fields of computer vision, and natural language processing.
These models seek to approximate complex relationships between input data and a set of corresponding labels.
Bayesian neural networks differ from these conventional neural networks by including some feature that captures both the uncertainty about the data -- aleatoric uncertainty -- and uncertainty about the parameters -- epistemic uncertainty.

Work by Gal and Ghahramani~\cite{gal2016dropout} demonstrated that a certain use of a common deep learning regularization technique, dropout, can approximate a spike and slab Bayesian prior.
Dropout is a technique where by setting a hyperparameter known as the dropout rate, the neuron will output zero with that probability~\cite{srivastava2014dropout}.
The work of Gal and Ghahramani applied a high probability dropout to each layer of the network to approximate Bayesian inference, capturing the aforementioned aleatoric and epistemic uncertainty.
Prior work by the author~\cite{galinkin2021influence} demonstrates tension between this sort of Bayesian dropout and differential privacy, suggesting that some amount of information may inadvertently be ``leaked'' by the use of uncertainty quantification techniques.

Expressing uncertainty provides useful information -- in many cases, a highly confident prediction suggests that the network has seen similar samples in training.
This benefit to AI safety, in making predictions on out-of-distribution data less confident, can also provide extremely valuable information when conducting membership inference attacks.
Membership inference attacks against machine learning models seek to determine whether or not a data sample was in the training data for the target.
Shokri \textit{et al.}~\cite{shokri2017membership} developed a method for conducting these attacks against machine learning models, conducting closed-box queries that return only the output probabilities for each class to predict whether or not the sample input to the model was in the training data set.
This was done by first training $k$ so-called ``shadow models'' that imitate the behavior of the target, trained on a synthetic dataset that is from approximately the same distribution as the target, where $k$ is the number of output classes.
Then an attack model is trained for each of the $k$ models, taking the outputs of the $k$ models and making a prediction of whether or not the sample was in or out of the training data.

Salem \textit{et al.}~\cite{salem2019ml} refined Shokri~\textit{et al.}'s procedure, training only a single shadow model, and making a number of other improvements that improve efficacy and reduce both training time and the amount of synthetic training data.
The methods used in our paper are derived from these refinements.
Interestingly, Salem \textit{et al.} recommend dropout as a potential defense against membership inference, albeit not enough to provide the approximate Bayesianism of Gal and Ghahramani's methods.

Adversarial examples -- specially crafted samples which seek to perturb images in imperceptible ways to cause misclassification -- were introduced to the world by Szegedy~\textit{et al.}~\cite{szegedy2013intriguing} as a peculiar feature of differentiable models.
Further work by Carlini and Wagner~\cite{carlini2017towards} expanded the understanding and evaluation of these adversarial examples, and considered the potential threat models associated with adversarial examples.
This same work by Carlini and Wagner also introduced three new attack algorithms that each minimally distort images under a given distance metric.
Work by Suciu~\textit{et al.}~\cite{suciu2019exploring} considers the specific question of adversarial examples against deep learning-based malware detectors, highlighting the potential for more sophisticated detection evasion techniques using adversarial machine learning.

Our work is thus further motivated by the assertion of Smith and Gal~\cite{gal2018sufficient} that Bayesian neural networks cannot have adversarial examples under some set of conditions.
In this work, the authors develop a strong theoretical justification that under idealized circumstances, Bayesian networks will capture perfect epistemic uncertainty and thus cannot have adversarial examples.
To exercise this theory in an empirical way, they generate a synthetic dataset with complete knowledge of the ground-truth image-space density and find that for Hamiltonian Monte Carlo -- the gold standard for Bayesian Neural Network inference -- adversarial crafting fails.
Our work instead considers the susceptibility of Bayesian models to adversarial examples on familiar benchmark datasets compared with a baseline model rather than synthetic datasets.

In general, adversarial attacks work via the same principle -- given a neural network with fixed parameters, $F(x) = y$, we define the classifier $C(x) = \arg \max_i F(x)_i$ to output a label for the input $x$.
Let $C^*(x)$ be the correct label of $x$.
Then, given a valid input $x$ and a target class $t \neq C^*(X)$, we search for a small perturbation $\delta$ such that for distance metric $D$,
\[
\begin{split}
\text{minimize } D(x, x + \delta) \\
\text{such that } C(x + \delta) = t
\end{split}
\]
The distance metric $D$ is most often some $\ell_p$ norm, with $p$ most often set to 0, 2, or $\infty$.
In untargeted cases, we search instead for some valid input $x'$ which meets the same minimization criteria and constraints as $x + \delta$.
This approach, however, is less common than generating noise or a patch to be applied.

\section{Methods}
To assess the effect of Bayesianism on the efficacy of membership inference attacks and adversarial examples, we leverage a total of four models.
Each model architecture features a standard, dropout-less control and a Bayesian experimental model.
All of the models are trained and tested on both the MNIST~\cite{lecun1998mnist} and the CIFAR-10~\cite{krizhevsky2009learning} datasets, two well-researched benchmarks.
This provides a good basis for comparison to other model architectures and allows us to draw more specific conclusions about the effect of Bayesianism absent variables around the data and architecture.
All of the models and experiments were conducted on a single server, with a single RTX 2080 Ti, 128 GB of RAM, and an AMD Threadripper 1950x.

\subsection{Models} \label{sec:models}
Our first experimental architecture is based on the LeNet-5~\cite{lecun2015lenet} architecture, which features 5 hidden layers.
The second experimental architecture is based on the ResNet-18~\cite{he2016deep} architecture that takes advantage of residual connections to avoid the vanishing gradient problem -- a problem which plagues very deep, non-residual networks.
Although our datasets are quite simple, these two architectures are widely studied and easy to train given computational limitations.
In order to make the aforementioned models Bayesian, our work leverages the spike and slab prior of Gal and Ghahramani~\cite{gal2016dropout} within the experimental models.
This ensures the highest degree of similarity between the Bayesian and non-Bayesian architectures in order to isolate the effect of Bayesianism.
All of the models were trained with early stopping to prevent overfitting, but no hyperparameters were tuned in training the architectures to assure the highest degree of similarity between Bayesian and Non-Bayesian models.

\subsection{Attacks}
Attacks on machine learning systems come in many forms, from more conventional denial of service or code execution-style attacks on the libraries~\cite{stevens2017summoning} to the ML-specific denial of service via sponge examples~\cite{shumailov2020sponge}, there are many facets where machine learning systems can be exploited.
We consider two classes of attacks from this array: membership inference attacks and adversarial examples.
These two attack classes have both been well-studied, and target wildly different parts of the process -- membership inference seeking to attack the confidentiality of training data, and adversarial examles seeking to attack the integrity of the predictions.
Background on these is provided in Section~\ref{sec:background} and we detail the methods used in our experiments below.

\subsubsection{Membership Inference} \label{sec:mi_attack}
\begin{figure}[h]
\centering
\includegraphics[width=0.45\textwidth]{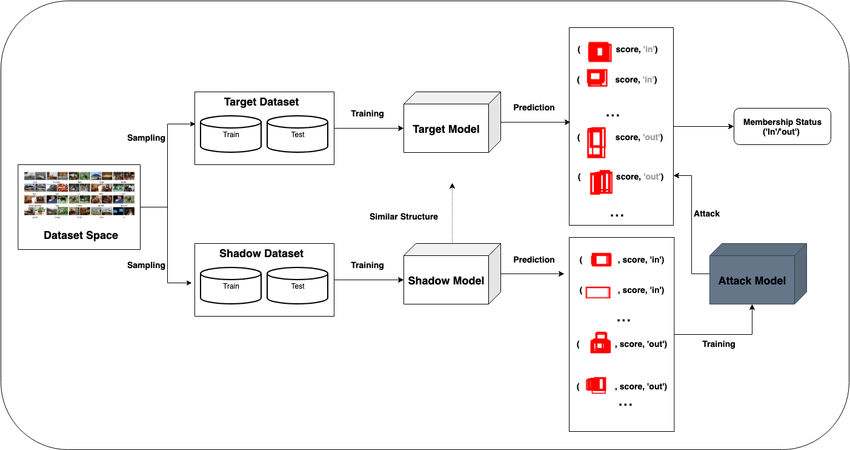}
\caption{Diagram of Membership Inference Attack} \label{fig:shadow}
\end{figure}

We leverage the prior work done by Salem \textit{et al.}~\cite{salem2019ml} on closed-box membership inference attacks.
This involves the creation of a synthetic ``shadow'' dataset with a similar distribution to the training distribution of the target model.
The shadow dataset is used to train a shadow model that approximates predictions for the target classifier.
Additional synthetic examples are generated and added to the shadow dataset, to be used as a validation dataset on which the classifier is not trained.
The shadow model and shadow dataset is then used to generate a set of output vectors from the classifier  and assigned an ``in'' or ``out'' label depending on whether or not the model was trained on the sample.
This in-or-out dataset is used to train a binary classifier to attack the model.
This process is visualized in Figure~\ref{fig:shadow}.
Further technical details are available in the paper by Salem \textit{et al.}

\subsubsection{Adversarial Examples}
Using the Microsoft Counterfit framework\footnote{https://github.com/azure/Counterfit}, we evaluate our models using two different attacks: a boundary attack~\cite{brown2017boundary} and the hop-skip-jump attack~\cite{chen2020hopskipjumpattack}.
These attacks are both closed-box attacks -- attacks which do not require direct access to the model and instead consider only inputs to and outputs from the model -- making the attack scenario more realistic than open-box models where all details are available to attackers.
Each model was targeted using both attacks over 100 iterations per attack, model, and dataset combination.
We evaluate attack efficacy as a percentage equal to the number of successful attack attempts.

The boundary attack is a decision-based attack which leverages an adversarial patching technique intended to cause misclassification in the real world.
It starts with a perturbation of the original image that is already adversarial -- classified as part of the target class -- and iteratively reduces the $\ell_2$ distance between the original image and the adversarial example by randomly walking along the boundary between the adversarial and non-adversarial region of the classifier.
This is done by first ensuring the adversarial sample $x'$ is in the input domain, in this case, a valid image.
Then, at step $k$, the perturbation $\eta^k$, should have a relative size of $\delta$:
\[\|\eta^k\|_2 = \delta \cdot D(x, x')\]
where $D$ is the distance metric being used.
In this case, we set $D(x, x') = \|x - x'\|_2^2$.
The value of $\eta$ is drawn from a Gaussian distribution, than rescaled and clipped to ensure the above criteria hold.
In any case where perturbing $x'$ by $\eta$ moves the adversarial sample out of the target class, the perturbation is discarded.

The hop-skip-jump attack is inspired by the boundary attack, but rather than rejecting perturbations that move the adversarial sample out of the target class, those perturbations are used to estimate the gradient direction and extends the attack to the $\ell_{\infty} norm$.
The attack is also both very query efficient, making it useful for evaluating real-world security of models, and comparable in efficacy to open-box attacks like Carlini-Wagner~\cite{carlini2017towards}.
Warde-Farley and Goodfellow~\cite{warde2016adversarial} have put forth the argument that $\ell_{\infty}$ is the optimal distance metric for adversarial examples, and so that is the distance metric we use in our implementation.

\begin{table*}[ht]
\centering
\begin{tabular}{@{}l|llll@{}}
\toprule
\textbf{Dataset and Model}        & \textbf{Test Accuracy}     & \textbf{Hop-Skip-Jump}       & \textbf{Boundary}       & \textbf{Membership Inference}\\ \hline
MNIST LeNet-5                   & 98.48\%           & 58\%                 & 52\%           & 63.91\% \\
Bayesian MNIST LeNet-5      & 98.01\%           & 52\%                 & 50\%           & 64.72\% \\
MNIST ResNet-18                 & 98.66\%           & 64\%                 & \textbf{70\%}           & 62.49\% \\
Bayesian MNIST ResNet-18    & 98.02\%           & \textbf{66\%}                 & 62\%           & \textbf{73.38\%}\\ \hline
CIFAR-10 LeNet-5                & 58.43\%           & \textbf{94\%}                 & \textbf{92\%}           & 58.86\% \\
Bayesian CIFAR-10 LeNet-5   & 58.56\%           & 85\%                 & 87\%           & 66.85\% \\
CIFAR-10 ResNet-18              & 75.40\%           & 69\%                 & 71\%           & 72.10\% \\
Bayesian CIFAR-10 ResNet-18 & 73.20\%           & 60\%                 & 57\%           & \textbf{77.16\%} \\ \hline
\bottomrule
\end{tabular}
\caption{Summary of results for test set accuracy, efficacy of adversarial attacks, and accuracy of membership inference against the model. Results for attacks with the highest efficacy on the dataset are in bold.} \label{tab:results}
\end{table*}

\section{Results}
The results in Table~\ref{tab:results} show that in all cases where a model is more Bayesian, it features a higher susceptibility to membership inference attacks compared with its non-Bayesian counterpart.
This strongly implies that a more Bayesian model is inherently more susceptible to membership inference, which is likely an artifact of the additional uncertainty captured on out-of-distribution samples.
When a Bayesian model makes an inference on a sample it has been trained on, is is substantially more confident than compared with samples that it has not seen, which lends itself to the efficacy of the attacks on these models.

\begin{figure}[h]
\centering
\includegraphics[width=0.4\textwidth]{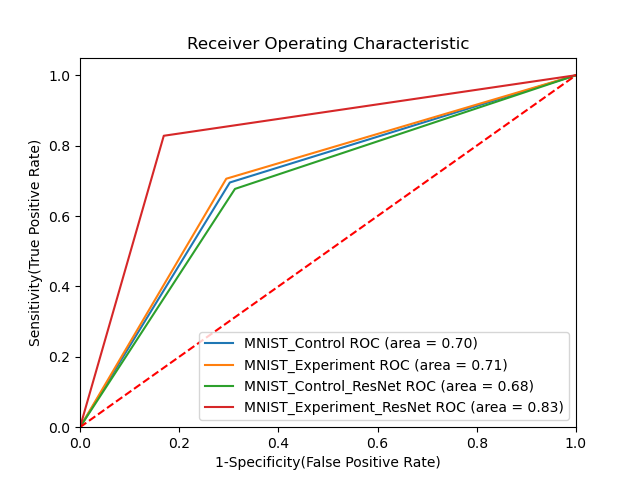}
\caption{Area under the curve plot for membership inference against MNIST trained model} \label{fig:auc_mnist}
\end{figure}

\begin{figure}[h]
\centering
\includegraphics[width=0.4\textwidth]{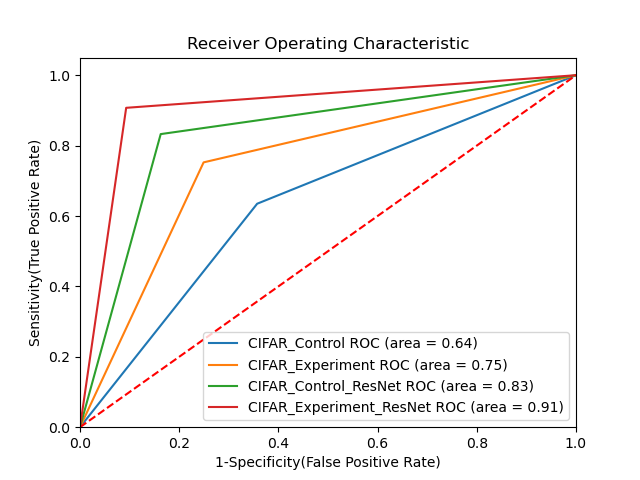}
\caption{Area under the curve plot for membership inference against CIFAR-10 trained model} \label{fig:auc_cifar}
\end{figure}

As we can observe from Figure~\ref{fig:auc_mnist}, at all points for models trained on MNIST, the ResNet with dropout is far more vulnerable to the membership inference attack than the other 3 models, which are clustered more closely together.
The curve is an artifact of the attack model described in Section~\ref{sec:mi_attack}, comparing the true positive and false positive rate.
In this case, a true positive is a correct identification that an observed sample is from the training dataset and a false positive is declaring the sample as part of the training dataset when it is not.
The closeness of the curves other than the Bayesian ResNet suggests that there may be too much noise in the predictions.
This issue could potentially be resolved with hyperparameter tuning.

The models trained on CIFAR-10, shown in Figure~\ref{fig:auc_cifar} show a greater variability in vulnerability to membership inference attacks than those trained on MNIST, though the AUC for ResNet with dropout still greatly exceeds the other three models.
We note, however, that no model trained on any dataset is especially close to the baseline AUC of 0.5, with the LeNet-5 trained on CIFAR-10 having the smallest AUC, 0.64.

\begin{table}[h]
\begin{tabular}{@{}l|llll@{}}
\toprule
\textbf{Dataset}        &\textbf{Bayesian}  & \textbf{Non-Bayesian}\\ \hline
MNIST                   & 69.05\%           & 63.2\%\\
CIFAR-10                & 72.01\%           & 65.48\% \\ \hline
\bottomrule
\end{tabular}
\caption{Membership inference efficacy between Bayesian and Non-Bayesian models per dataset} \label{tab:mi_detail}
\end{table}

Table~\ref{tab:mi_detail} breaks down the results from Table~\ref{tab:results} in detail, showing that in general, Bayesian models are approximately 10\% more vulnerable to membership inference attacks than their non-Bayesian counterparts independent of dataset.
This suggests that there is an inherent trade-off being made between quantifying uncertainty and protecting the confidentiality of a model's training data.
We discuss potential mitigations that may alleviate this trade-off in Section~\ref{sec:conclusion}.

In contrast with the membership inference results, Bayesian and non-Bayesian models proved approximately equally vulnerable to adversarial examples.
Efficacy of attacks demonstrated little correlation between model and dataset combination, and no one model type clearly more vulnerable the others across datasets.
On CIFAR-10, the non-Bayesian LeNet-5 proved more vulnerable to adversarial examples for both attacks, but the same model architecture was the second-most robust on average for the MNIST dataset.
The presence of Bayesianism seems to have no effect on the efficacy of adversarial examples in general.
This is reasonable given the fact that the decision boundaries for the various models and datasets should be quite similar and the attack methods used have no knowledge of the models inner workings, and must rely entirely on gradient estimation techniques.

\begin{table}[h]
\begin{tabular}{@{}l|llll@{}}
\toprule
\textbf{Dataset}     & \textbf{Hop-Skip-Jump}   &\textbf{Boundary} \\ \hline
MNIST                & 60\%                     & 58.5\% \\
CIFAR-10             & 77\%                     & 76.75\% \\ \hline
\bottomrule
\end{tabular}
\caption{Average efficacy of adversarial attacks across all models per dataset} \label{tab:adversarial}
\end{table}

Interestingly, the training dataset seems to have a much stronger relationship to the efficacy of adversarial samples than any model-specific artifact, as we can observe from Table~\ref{tab:adversarial}.
This is likely due to the additional complexity of the sample space for CIFAR-10, as compared with MNIST.
This follows from the sensitivity to well-generalizing features in the data, as described by Ilyas~\textit{et al}~\cite{ilyas2019adversarial}.
Since CIFAR-10 images are both larger and are color images, as compared with MNIST's black and white images, they have a comparative wealth of features that may be highly predictive but not robust.

\section{Conclusion} \label{sec:conclusion}
The results of our experiment are very clear on the question of membership inference attacks -- Bayesian neural networks are substantially more vulnerable to these types of attacks.
In sensitive use cases of artificial intelligence, we will want to have both uncertainty quantification and data privacy.
Unfortunately, these experiments suggest that we likely cannot have both.
Returning only the top-1 result with a probability may help alleviate this, as it is quite difficult to build a membership inference attack off of so little data.
However, this may impact the usefulness of the model in cases where humans need to make decisions based upon its output.

Luckily, it seems that in cases where adversarial examples are part of our threat model, Bayesianism does not increase nor reduce the opportunity for attacks to be successful.
Across the two data sets, the hop-skip-jump attack was most effective on alternatively the largest Bayesian model and the smallest non-Bayesian model while the boundary attack proved effective on two different non-Bayesian models across the two data sets.
The results of our experiment suggest that the most significant factor in the success of adversarial examples is the complexity of the dataset -- models trained on datasets with more space to perturb the image are far more vulnerable.
This suggests that Bayesianism should not affect your risk calculus in cases where adversarial examples are a threat.
In many cases where Bayesianism is being implemented, confidence on unexpected samples should be low, so there may even be other advantages against adversarial examples.

In future work, we seek to see how these attacks work on non-image data, such as text.
Privacy of text is more acutely relevant for most use cases, and prior work~\cite{carlini2020extracting} suggests that stronger forms of training data recovery are possible on these sorts of models.
Coupling that work with the increased complexity in the adversarial example space and recent breakthroughs~\cite{boucher2021bad} in adversarial examples on text classifiers make this realm a natural extension of the current work.
Additionally, future work might consider how certain defenses against membership inference and adversarial examplesperform when paired with Bayesianism.

\begin{acks}
Funded by the Auerbach Berger Chair in Cybersecurity held by Spiros Mancoridis, at Drexel University
\end{acks}

\bibliographystyle{ACM-Reference-Format}
\bibliography{references}

\end{document}